\documentclass[draft=True, sigplan, nonacm]{acmart}

\AtBeginDocument{%
  \providecommand\BibTeX{{%
    \normalfont B\kern-0.5em{\scshape i\kern-0.25em b}\kern-0.8em\TeX}}}

\setcopyright{iw3c2w3}
\copyrightyear{2019}

\begin{document}

\title{Pattern Spotting in Historical Documents Using Convolutional Models}


\author{Ignacio \'Ubeda}
\email{ignacio.ubeda@ing.uchile.cl}
\affiliation{%
  \institution{University of Chile}
  \streetaddress{Beauchef 850, Santiago}
  \city{Santiago}
  \country{Chile}
}

\author{Jose M. Saavedra}
\email{jose.saavedra@orand.cl}
\affiliation{%
  \institution{Computer Vision Group, Orand S.A.}
  \streetaddress{Estado 360}
  \city{Santiago}
  \country{Chile}
}

\author{St\'ephane Nicolas}
\email{Stephane.Nicolas@univ-rouen.fr}
\author{Caroline Petitjean}
\email{Caroline.Petitjean@univ-rouen.fr}
\author{Laurent Heutte}
\email{Laurent.Heutte@univ-rouen.fr}
\affiliation{%
  \institution{Normandie Univ, UNIROUEN, UNIHAVRE, INSA Rouen, LITIS}
  \streetaddress{Technopole du Madrillet, Avenue de l'Universit\'e, 76800 Saint-Étienne-du-Rouvray}
  \city{Rouen}
  \country{France}
}



\begin{abstract}
  Pattern spotting consists of searching in a collection of historical document images for occurrences of a graphical object using an image query. Contrary to object detection, no prior information nor predefined class is given about the query so training a model of the object is not feasible. In this paper, a convolutional neural network approach is proposed to tackle this problem. We use RetinaNet as a feature extractor to obtain multiscale embeddings of the regions of the documents and also for the queries. Experiments conducted on the DocExplore dataset show that our proposal is better at locating patterns and requires less storage for indexing images than the state-of-the-art system, but fails at retrieving some pages containing multiple instances of the query. 
\end{abstract}





\maketitle

\section{Introduction}\label{sec:introduction}
Today, digitalization of historical document collections helps grant access to more users while limiting contact with the real materials. For historians, there is a need for automated software tool that will allow them to establish correspondences between documents or parts of documents, whether on textual content or on graphical parts \cite{yarlagadda:ACCV:2010}. Thanks to the spectacular advances in the field of computer vision and machine learning, it is now possible for computers to analyze very large volumes of images and find matchings within a few seconds.

Historical documents contain mainly text, but also some graphical patterns or objects. For the recognition of textual content, in both printed and handwritten documents, recognition techniques (OCR and HWR) have made great progress, especially thanks to deep learning techniques (CNN, RNN, BLSTM), resulting in a great amount of publications. Nevertheless handwriting recognition (HWR) remains a complicated task especially for historical documents. For this reason, in an information retrieval perspective, word spotting is an interesting alternative to full text recognition, raising interest especially in recent years \cite{Giotis2017}. But there is little work on the automatic analysis of the graphical content of historical documents. There are content based information retrieval systems to search similar images in photo databases, but no tool to locate similar patterns in historical documents or in art collections.
Computers can solve these tasks with quite good results (image retrieval) but precise localization of small patterns is still a challenging task. In \cite{yarlagadda:ACCV:2010} the authors explore the question of category level object detection, for semantic based indexing, in the context of a benchmark dataset for cultural heritage studies. For that they propose a benchmark image dataset of medieval images with groundtruth information and a detection system that accurately localizes objects.  In \cite{en2016new, En2016PR}, En et al. have proposed a benchmark dataset, called "DocExplore", which is publicly available online \cite{psWebsite}, and a complete system able to perform image retrieval, using query by example paradigm, and pattern spotting tasks. By "pattern spotting" we mean the task of locating as precisely as possible, in the images of an indexed database of document images, the different instances of a given object, i.e. an image query. This queried image is generally smaller than the images indexed in the database, which makes this task difficult, in addition to problems of representation variability. The pattern spotting task we consider is quite similar to the near-duplicate figures searching presented in \cite{Rakthanmanon2011}, in the same context of historical document indexing.


Whereas previous work was based on classical approaches, in this paper we propose to build upon the method presented in \cite{En2016PR}  and to rely on deep learning techniques for feature extraction. In Sec. \ref{sec:related_work} we motivate our choice by presenting related work, and we detail our proposal in Sec. \ref{sec:methodology}. We explain in Sec. \ref{sec:results} the adopted protocol to evaluate our system and discuss the results. Finally we conclude this paper in Sec. \ref{sec:conclusion} by proposing future work.

\section{Related Work}\label{sec:related_work}

In \cite{En2016PR}, a complete system for searching images and locating small graphic objects in images of medieval documents has been proposed. This system is based on a first extraction/indexing of regions of interest in the image (region proposal / Binarized Normed Gradients), a characterization of these regions by ad-hoc descriptors (Vector of Locally Aggregated Descriptors and Fisher Vector), and a query similarity search integrating compression and approximation techniques (Inverted File, Product Quantization and Asymmetric Distance Computation). While this system has shown good performance on the corpus of document images of interest \cite{en2016new}, it suffers from a number of weaknesses that make this system unsuitable for other types of document images (colour information is not currently used, for example), sensitive to variations in size, shape, colour and more generally, style of the patterns to be detected. On the other hand, this system does not easily support scaling up and requires post-processing for fine localization of objects in regions of interest using, for example, traditional matching methods.
These last years, deep learning techniques have been successfully used, in various tasks, especially convolutional neural network (CNN) for image analysis and recognition. Content-based image retrieval is no exception, in which deep features and deep hashing are largely used in several ways \cite{wan2014deep}, and also word spotting, with deep architectures such as PHOCnet \cite{Sudholt2016}. In recent work, style recognition \cite{lecoutre2017recognizing} and visual pattern discovery \cite{shen2019discovery} in art collections have also been considered using deep learning approaches with interesting results.

\section{Our Deep Proposal}\label{sec:methodology}

\begin{figure*}
  \centering
  \includegraphics[width=0.9\linewidth]{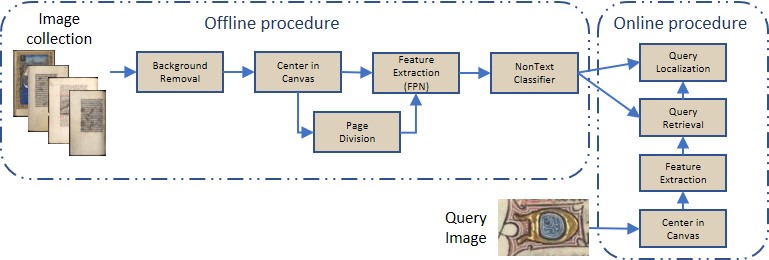}
  \caption{Overview of the pipeline with the offline and online procedures.}
  \Description{Overview of the system pipeline}
  \label{fig:overview}
\end{figure*}

Our proposal is composed of two stages: one working in an offline manner and the other in an online one.  The offline stage is focused on processing the historical documents including tasks like feature extraction and indexing, while the online stage is focused on processing the input query. The second stage includes tasks like query feature extraction and searching in the document features. Both stages are based on convolutional models. An overview of the pipeline is given in Fig. \ref{fig:overview}. Hereafter, we detail each processing step.


\subsection{Background Removal}

The goal of this \textit{preprocessing} stage is to keep only the informative regions of each page\footnote{Here an ``informative region'' is a region of the page which may contain symbols and/or handwriting text but not background.}. The method is based on a region growing paradigm starting from the center of the page, similar to that explained in \cite{En2016PR} with two main modifications. First, we add morphological operations (erosion and dilation) to improve the image binarization and second, we modify the stopping criterion: pixel borders are added until background pixels of the considered border zone exceed 95\% (i.e. almost all of the pixels are background pixels). 


\subsection{Center in Canvas and Page Division}

We need that the shape of the CNN input be fixed. Most works face this problem by resizing the input images to a fixed size. The problem with this approach is that small instances of queries could be taking out since small details of image regions could be lost after resizing.

\begin{figure}[ht]
  \centering
  \includegraphics[width=0.9\linewidth]{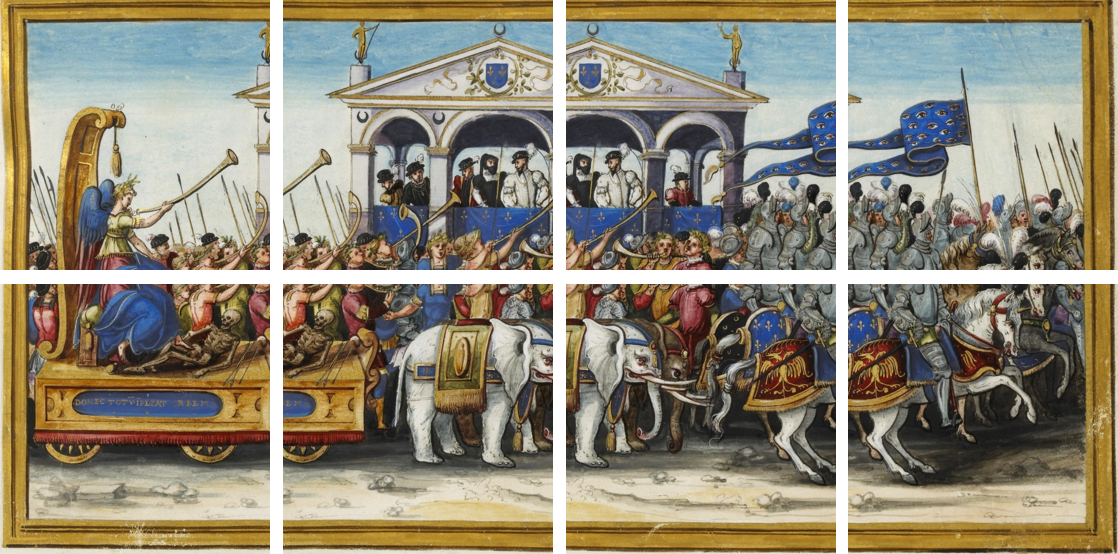}
  \caption{Sub-pages example. All sub-pages have the same shape ($10^3 \times 10^3$); overlapping may exist between them.}
  \Description{Sub-Pages example}
  \label{fig:page_division_subpage}
\end{figure}

To address this problem, we use a simple approach in which the page is centered in a black canvas to fix the required input size ($10^3 \times 10^3$). However, there are also pages that exceed the $10^3$ size in one (or both) dimension(s). For this case, we divide the page through the following procedure: first, we locate the corner centers matching a $10^3 \times 10^3$ template in each corner. Then, we create an equally spaced grid with those corner centers and finally we match a template of the desired shape in each center of this grid. In Fig. \ref{fig:page_division_subpage} we show the output sub-pages for an example page greater than $10^3$ in both dimensions. As noted in Fig. \ref{fig:overview}, we apply the same procedure for the queries to fix the input shape but instead of a black canvas, we use a historical "blank" page (i.e. the background page from a manuscript) as a canvas to simulate the special textures that these types of pages have. We discuss the impact of this  modification in Sec. \ref{sec:results}.



\subsection{Feature Extraction}

Let us recall the concept of Receptive Field before detailing this stage. The Receptive Field (RF), in the context of CNN, is defined as the local region of the input that affects a particular neuron (in ordinary words, what the neuron is ``looking at''). The RF size is tightly related with the kernel size used in the convolutional layers and, as most CNN architectures also do spatial pooling operations through the forward pass, deeper neurons have bigger RF than shallow neurons. For a detailed explanation of this concept please refer to \cite{luo2016understanding}.

\begin{figure}[ht]
  \centering
  \includegraphics[width=\linewidth]{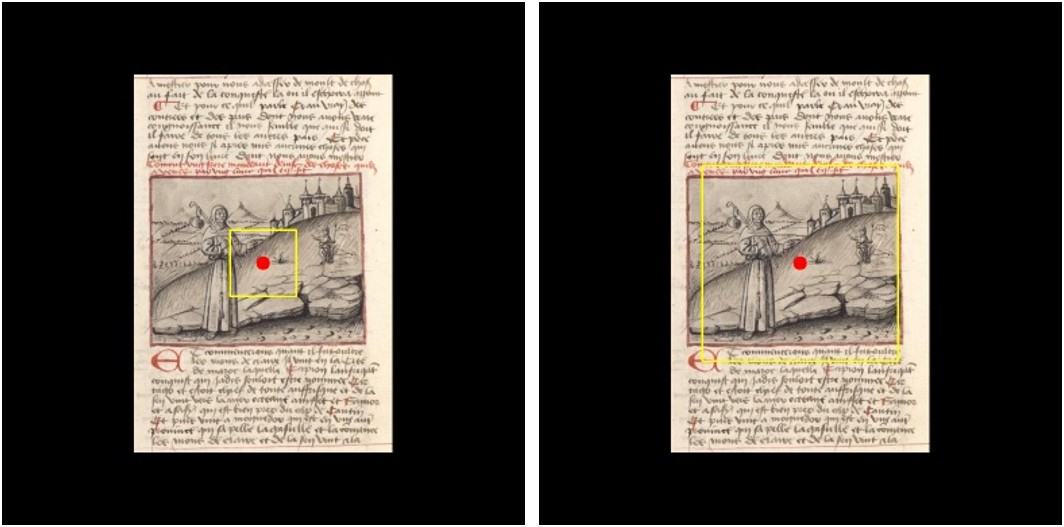}
  \caption{Neuron RF across pyramid levels example. The left corresponds to $P_3$ while the right to $P_4$. Note that deeper neurons have bigger RF than shallow neurons.}
  \Description{RF across levels}
  \label{fig:RF_across_levels}
\end{figure}

We use RetinaNet \cite{lin2018focal} as a feature extractor (for both, pages and queries) but only the feature pyramid net was kept (the ``head'' of the net, class and box subnets, was discarded). This allows to get features in a patch-based manner. So in a forward stage, we can get feature vectors for different patches on the same image. For each page, we extract $P_3$, $P_4$ and $P_5$ feature map levels to get embeddings of all the regions at different scales. Those embeddings are based on the RF of each neuron at each level, hence $P_3$ extracts embeddings for the smallest regions and $P_5$ for the biggest ones. The distance between two RF centers of two neighbor neurons at the corresponding feature map is 8, 16 and 32 pixels respectively for $P_3$, $P_4$ and $P_5$\footnote{As the smallest query size is $\approx$ 10x20 pixels, we expect that some neuron at $P_3$ level is ``looking at'' that query because the distance between two neighbors RF centers is smaller than the smallest query size.}. In Fig. \ref{fig:RF_across_levels} we draw the RF for one neuron at $P_3$ and $P_4$ for an example page.

Contrary to the pages where we extract a pyramid of feature maps, for the queries we only extract one embedding per each query. As these are centered in the input, we keep the center neuron at level $P_k$ as the embedding for each one. We assign a query of width $w$ and height $h$ to the level $P_k$ by Eq. \ref{eq:select_pyramid_level} and we set $k_0=4$ identically as in \cite{lin2017feature}. The depth of each level of our pyramid is $256$, hence we have embeddings for queries and for - all the regions at three different scales of - pages as $256$D feature vectors.

\begin{equation}
    k= \left \lfloor k_0 + \log_2(\sqrt{wh}/224) \right \rfloor
    \label{eq:select_pyramid_level}
\end{equation}

Note that RetinaNet was not trained on the DocExplore dataset; instead, it was trained on COCO dataset \cite{lin2014microsoft} and we are using it only as a feature extractor with the same weights adjusted by COCO.

\subsection{NonText Classifier}

\begin{figure}[ht]
  \centering
  \includegraphics[width=0.5\linewidth]{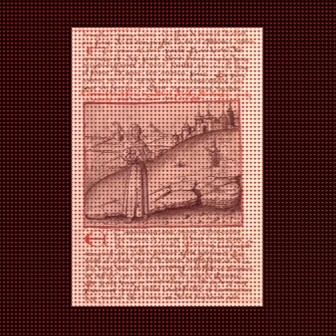}
  \caption{RF centers (red points) for all neurons at $P_3$ level}
  \Description{RF center P3}
  \label{fig:RF_centers_P3}
\end{figure}

With the previous stage, we have embeddings for each query and each page. So we could do an exhaustive search on each page with all the neurons at a given level for a query retrieval, but it would be computationally expensive. In the example of Fig. \ref{fig:RF_centers_P3}, we have drawn the center of each RF neuron for the $P_3$ level. For our fixed input, $P_3$ level has a spatial resolution of $125\times125$ which means that for each page, a dense search will compare $125^2$ feature vectors with the embedding of the query. That is infeasible if we want an online retrieval system. Also, as can be seen in Fig \ref{fig:RF_centers_P3}, many of the RF centers are located in the - artificially created - black canvas (i.e. we have many embeddings only of black pixels) and we know for sure that the query will never appear in those regions. We can address this problem training a NonText classifier as a simple ``region proposal'' step.

To create the train set we sampled (79 of 1500) pages and manually labeled them with bounding box of NonText ROI. Then, we assigned each neuron embedding to a label based on the intersection with each ROI bounding box (i.e. we used the same $256$D feature vector provided by RetinaNet to train our NonText classifier) at each pyramid level and we split the sets in train/validation/test with proportions 0.6/0.25/0.15, respectively. The size of each set depends of the considered pyramid level. Let $R_k$ be one dimension of the spatial resolution of the pyramid level $k$ (e.g. $R_k=125$ for level $P_3$) and $p_s$ the proportion of the set $s$, then the size of the set $s$ at level $k$, ${size}^k_s$, is given by $size^k_s = 79 \times R_k^2 \times p_s$.  

\begin{figure}[ht]
  \centering
  \includegraphics[width=0.5\linewidth]{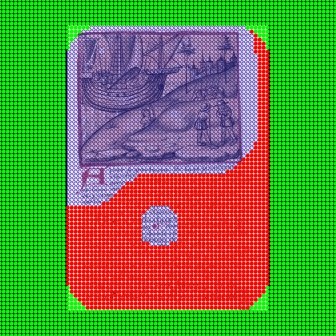}
  \caption{Example of a label page at level $P_3$ for the NonText classifier. Each point is a RF center of a neuron. Green is for \textit{black}, red for \textit{text} and blue for \textit{non-text} class.}
  \Description{NonText classifier}
  \label{fig:NonText_clf}
\end{figure}

In Fig. \ref{fig:NonText_clf} we show an example of a label page at level $P_3$. Three classes are considered: \textit{black} (for the black canvas pixels), \textit{text} and \textit{non-text}. Finally, we trained a Random Forest classifier at each level and we use it to predict in all pages. Batches of embeddings were created with only the \textit{non-text} predicted class. 

\subsection{Query Retrieval and Query Localization}

For retrieval, queries are searched only at the same level they were assigned to (e.g. if a query was assigned at $P_4$ level, then it is looked for only at the $P_4$ level of the pages). We use \emph{dot distance} as a similarity measure. This procedure finds the RF center of the closest embeddings to the query embedding. However, as the pages are cropped (by the background removal stage), centered in the black canvas and some of them were divided in sub-pages, we translate the coordinates of the RF center to the original page.

\begin{figure}[ht]
  \centering
  \includegraphics[width=0.9\linewidth]{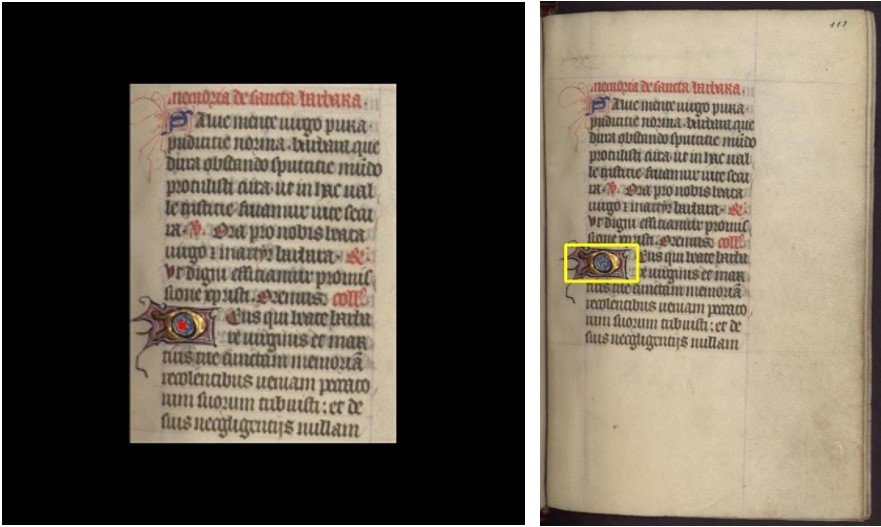}
  \caption{Bounding box localization for an instance of category ``D'' (letter ``D''  as a dropped initial). The red point in the left image corresponds to the RF center of the closest embedding while the right image shows the bounding box localization with the center already translated.}
  \Description{bounding box localization}
  \label{fig:BB_Localization}
\end{figure}

Finally, for the bounding box localization we simply center the query template at the found center. Fig. \ref{fig:BB_Localization} shows an example of this entire stage for category ``D''. Last but not least, we add a \textit{postprocessing} step where we discard the localization if the bounding box is not entirely contained in the original page (particularly useful for big queries) and do non-maximum suppression to avoid overlapping bounding box retrievals for the same instance (useful for sub-pages).

\section{Experiments and Evaluation}\label{sec:results}

\subsection{Experimental Protocol}

We make use of the DocExplore dataset \cite{en2016new,psWebsite} to compare our proposal to the state-of-the-art approach (Sec. \ref{sec:related_work}). Three important challenges come up with this particular dataset: (1) queries and documents are hand-drawn, introducing intra-class variation in the query-query and query-page relation, (2) queries are generally small (as small as $\approx 10 \times 20$ pixels) and (3) images are noisy and commonly degraded because all manuscripts are dated back from 10th to 16th century. 1447 possible queries among 35 different categories and 1500 pages are used for the experimentation.

Two tasks, \textit{image retrieval} and \textit{pattern spotting}, are considered in the evaluation. The aim of the \textit{image retrieval} task is to retrieve pages that contain the given query, regardless of its position in the page. On the other hand, the \textit{pattern spotting} task takes one step further and localizes the query within the retrieved page (i.e. the system has to retrieve the page and the bounding box for the given query). Note that one page may contain multiple instances of the same query. In this case the \textit{pattern spotting} task will count all the instances of the same page as correct retrievals while the \textit{image retrieval} task will count the page only as one retrieval. Finally, mean Average Precision (mAP) is used as the evaluation metric for both tasks. For the \textit{pattern spotting} task, the IoU between the returned box and the ground truth must be superior to 0.5 to consider the retrieval as a correct one.  

\subsection{NonText Classifier Results}

Even though \textit{accuracy} is the common metric reported in classifier systems, we argue that recall (particularly of the \textit{non-text} class) is the most important metric in this case. As we are predicting NonText regions (where we are going to search the query later), a poor recall for the \textit{non-text} class would imply that we are missing potential NonText regions where the query may appear (``query regions''). To the contrary, if we consider the extreme case of recall equals to 1, it means that we are predicting all the NonText regions correctly and our region proposal method is working as expected. 



\begin{figure}[ht]
  \centering
  \includegraphics[width=0.9\linewidth]{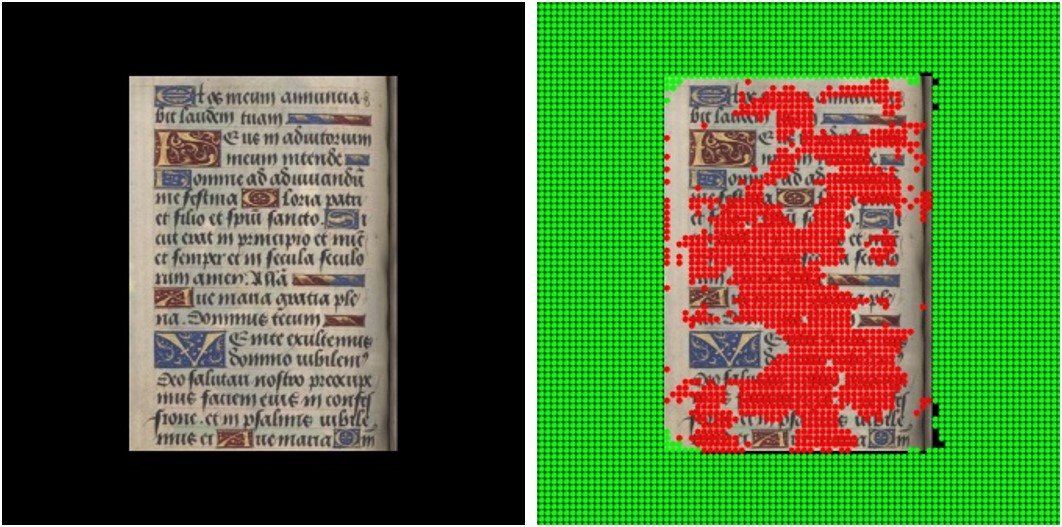}
  \caption{NonText classifier prediction example at level $P_3$. The \textit{non-text} class centers are not drawn.
  }
  \Description{NonText Classifier Predicted}
  \label{fig:non_text_clf_pred}
\end{figure}

The recall rates at each pyramid level for the test set are presented in Table \ref{tab:non_text_clf} where we also present the overall accuracy of the classifier only for completeness.
We can see from Table \ref{tab:non_text_clf} that as we go bottom up through the pyramid, results get worse. This is mainly because the RF size increases and classes are mixed in the area covered by the RF. Finally in Fig. \ref{fig:non_text_clf_pred} we show a page example of the regions predicted by the classifier.

\begin{table}
  \caption{NonText classifier results on the test sets.}
  \label{tab:non_text_clf}
  \begin{tabular}{crr}
    \toprule
    Pyramid Level & Recall & Accuracy\\
    \midrule
    $P_3$ & 0.997 & 0.975\\
    $P_4$ & 0.991 & 0.970\\
    $P_5$ & 0.985 & 0.953\\
  \bottomrule
\end{tabular}
\end{table}

\subsection{Image Retrieval and Pattern Spotting Results}

Results for both tasks for three different configurations are shown in Table \ref{tab:im_ps_results}. Dense configuration corresponds to an exhaustive search, i.e. without the NonText classifier. BlackCanv. and TemplCanv. correspond to the centering of the query in a black and an historical blank page canvas, respectively. One can first notice from Table \ref{tab:im_ps_results} that our NonText classifier is working as expected. No potential query region is missed and, thanks to the improvement in both tasks, we indeed are eliminating false positive regions. Despite the little improvement in mAP (from 0.286 to 0.300), the query search time is largely improved, switching from $\approx 3$ min. to $\approx 5$ sec. per query. Secondly, one has to notice the improvement made with the modification in the canvas. Whereas the black canvas introduces some noise, our historical blank page proposal allows us to simulate the textures of old manuscript pages and enables a 30\% improvement in the \textit{image retrieval} task.

\begin{table}
  \caption{mAP results for \textit{image retrieval} and \textit{pattern spotting} tasks for different configurations.}
  \label{tab:im_ps_results}
  \begin{tabular}{lrr}
    \toprule
    Configuration & \textit{image retrieval} & \textit{pattern spotting}\\
    \midrule
    Dense + BlackCanv. & 0.286 & 0.139\\
    NonText clf + BlackCanv. & 0.300 & 0.143\\
    NonText clf + TemplCanv. & 0.386 & 0.173\\
  \bottomrule
\end{tabular}
\end{table}

\begin{figure}[ht]
  \centering
  \includegraphics[width=0.9\linewidth]{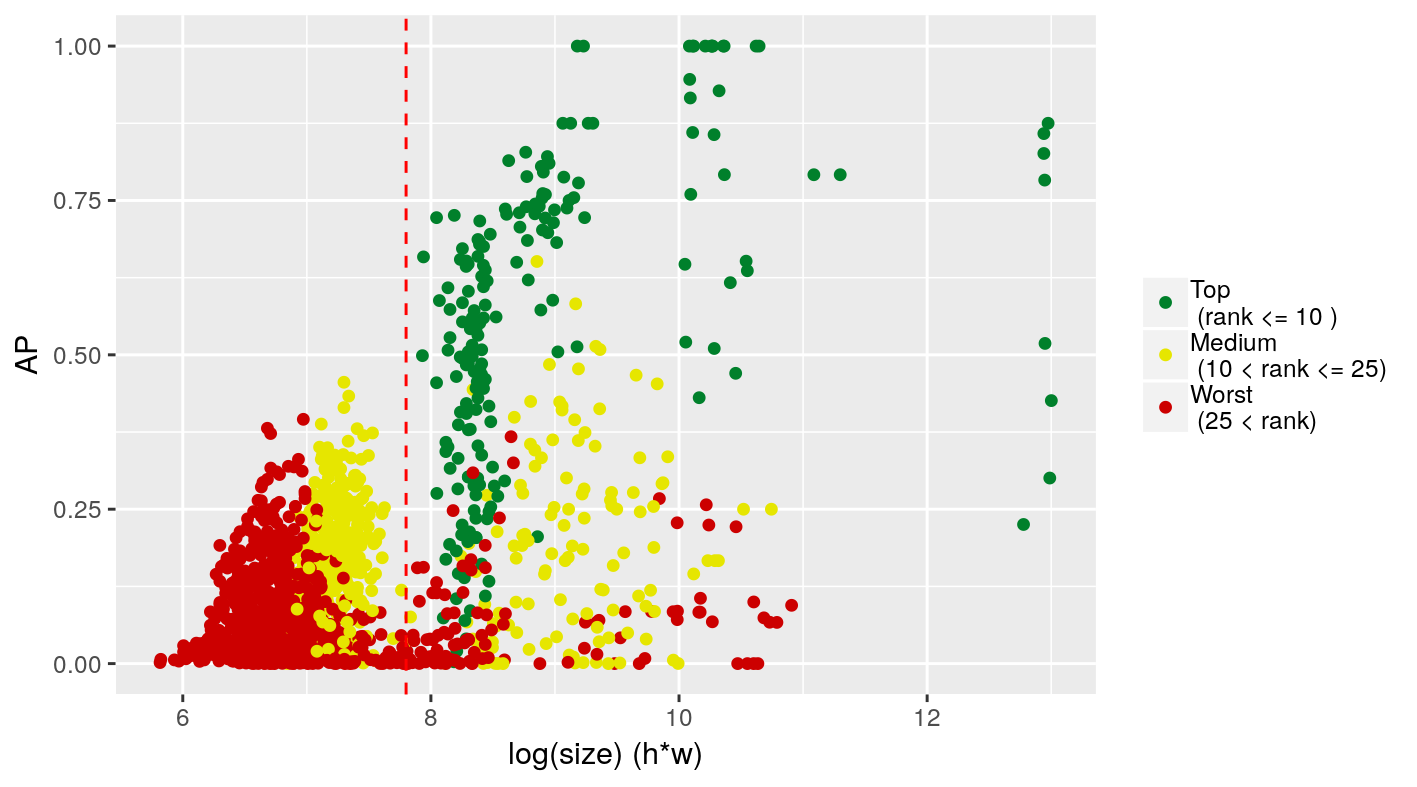}
  \caption{Average Precision (AP) against log size ($h \times w$) of the 1447 queries. Each color indicates the performance of the entire category to which the query belongs.}
  \Description{AP vs log size}
  \label{fig:AP_vs_logsize}
\end{figure}

In Fig. \ref{fig:AP_vs_logsize} we analyze the relationship between Average Precision (AP) and the query size. For this, we discretize the performance in three levels (Top, Medium, Worst) based on the category mAP ranking (e.g. a Top category is one that is in the first 10 positions in the ranking ordered by mAP). We see that our proposal also has poor performance with the smallest queries, as the state-of-the-art system \cite{En2016PR}. We think that the small size query problem could be tackled by considering a lower level of the pyramid, such as $P_2$, to reduce the RF size and let the embedding focus mostly on the query and not also on the canvas.

\begin{figure}[ht]
  \centering
  \includegraphics[width=0.9\linewidth]{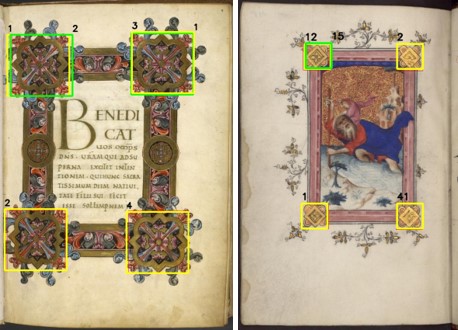}
  \caption{Localization example for top 50 retrievals for category ``Brace Ornament'' (left) and ``Corner Diamond'' (right). The ground truth is drawn in red, the bounding boxes returned by our system in yellow and by the state-of-the-art system in green. Ranking position of the bounding box retrieval is marked in top-left for our system and top-right for state-of-the-art system.}
  \Description{Localization comparison}
  \label{fig:loc_comp}
\end{figure}

We finally compare our proposal with the state-of-the-art results
\cite{en2016new} in Table \ref{tab:mAP_comparison}. First, as mentioned in Sec. \ref{sec:methodology}, our embeddings are 256D while the state-of-the-art system creates 4096D embeddings (so we reduce the necessary storage by a factor of 16 in the feature vectors). And second, we have an improvement of roughly 10\% in \textit{pattern spotting} task but our system is almost 30\% worse than the state-of-the-art system in \textit{image retrieval} task. That is contradictory to what one would expect, because an improvement of localization mAP should imply also an improvement of retrieval mAP. However, the latter is only true when we have one (and only one) query per retrieval. In this dataset, instances of the same category may appear on the same page several times and this explains why the \textit{pattern spotting} results improve while the \textit{image retrieval} ones get worse. Therefore, our proposal is better at locating patterns on the same page than locating them in multiple pages. This explanation is visually validated in the examples shown in Fig. \ref{fig:loc_comp}.   

\begin{table}
  \caption{mAP results for \textit{image retrieval} and \textit{pattern spotting} tasks comparison with state-of-the-art system.}
  \label{tab:mAP_comparison}
  \begin{tabular}{lrr}
    \toprule
    System & \textit{image retrieval} & \textit{pattern spotting}\\
    \midrule
    Our proposal & 0.386 & 0.173\\
    State-of-the-art \cite{en2016new} & 0.580 & 0.157\\
  \bottomrule
\end{tabular}
\end{table}

\section{Conclusions}\label{sec:conclusion}

In this work, a deep learning system is proposed for spotting and retrieving image patterns in historical documents. Compared with the state-of-the-art system on the DocExplore dataset, our system is better at locating patterns and requires less storage for indexing images, but fails at retrieving several pages where the query is contained. Both systems have poor localization performance with small queries, showing one of the main difficulties of spotting patterns precisely (size variability between and within the categories).

However, recall that we used a pre-trained net as feature extractor without fine-tuning for this particular type of images. Fine-tuning our feature extractor and considering a lower pyramid level ($P_2$) to search could improve the results (particularly for small queries) and this will be our guidelines of future work. 

\section*{Acknowledgement}
The authors would like to thank Sovann En for his great help and to CONICYT for funding this work througth the proyects  PFCHA/MAGISTER NACIONAL/2018 - 22180111 and Stic-Amsud 19-STIC-04.

\bibliographystyle{ACM-Reference-Format}
\bibliography{paper}

\end{document}